\documentclass[12pt]{article}

\usepackage{scicite}

\usepackage{times}
\usepackage{amsfonts}
\usepackage{graphicx}
\usepackage{amsmath}
\usepackage{color}
\usepackage{diagbox}

\newenvironment{sciabstract}{%
\begin{quote} \bf}
{\end{quote}}

\title{AES: Autonomous Excavator System for Real-World and Hazardous Environments} 

\author
{Jinxin Zhao$^{1}$, Pinxin Long$^{2}$, Liyang Wang$^{1}$, Lingfeng Qian$^{2}$, \\
Feixiang Lu$^{2}$, Xibin Song$^{2}$, Dinesh Manocha$^{3}$, Liangjun Zhang$^{1*}$\\
\\
\normalsize{$^{1}$Baidu USA, Sunnyvale, CA, 94089}\\
\normalsize{$^{2}$Baidu Research Institute, Beijing, China, 100085}\\
\normalsize{$^{3}$  University of Maryland, College Park, MD 20742}\\
\\
\normalsize{$^\ast$Corresponding Author. E-mail: liangjunzhang@baidu.com}
}

\date{}

\begin{document} 


\baselineskip24pt


\maketitle


\begin{sciabstract}
 Excavators are widely used for material-handling applications in unstructured environments, including mining and construction. 
 The size of the global market of excavators is $44.12$ Billion USD in 2018 and is predicted to grow to $63.14$ Billion USD by 2026~\cite{exanalysis}.
 Operating excavators in a real-world environment can be challenging due to extreme conditions and rock sliding, ground collapse, or exceeding dust. 
 Multiple fatalities and injuries occur each year during excavations\cite{afanuhpreventing}.
 An autonomous excavator that can substitute human operators in these hazardous environments would substantially lower the number of injuries and can improve the overall productivity.
In this paper, we present an autonomous excavator system (AES), for material-loading and soil excavation tasks. Our system is capable of handling different environments and uncertainties and uses a novel architecture that combines perception and planning. We use multiple perception sensors along with advanced object material and  texture detection algorithms. We present novel motion planning algorithms that combine learning-based techniques with optimization-based methods and are tightly coupled with the perception modules. We have evaluated the performance on compact and standard excavators in many complex indoor and outdoor scenarios corresponding to waste material handling, soil excavation, rock manipulation, etc. 
AES is the first autonomous excavator system that has been deployed for real-world operations for long periods and can operate robustly in challenging scenarios. 
We demonstrate that our novel architecture with integrated perception, planning and control components improves the efficiency and autonomously handle different scenarios, as compared with prior autonomous excavator systems. 
In practice, AES achieves $24$ HPI (Hours Per Intervention), i.e. the system can  continuously operate for $24$ hours without any human assistance. Moreover, the amount of  material handled by AES per hour is closely equivalent to that of an experienced human operator.
\end{sciabstract}

\section*{Introduction}

\begin{figure}
    \centering
    \includegraphics[width=115mm]{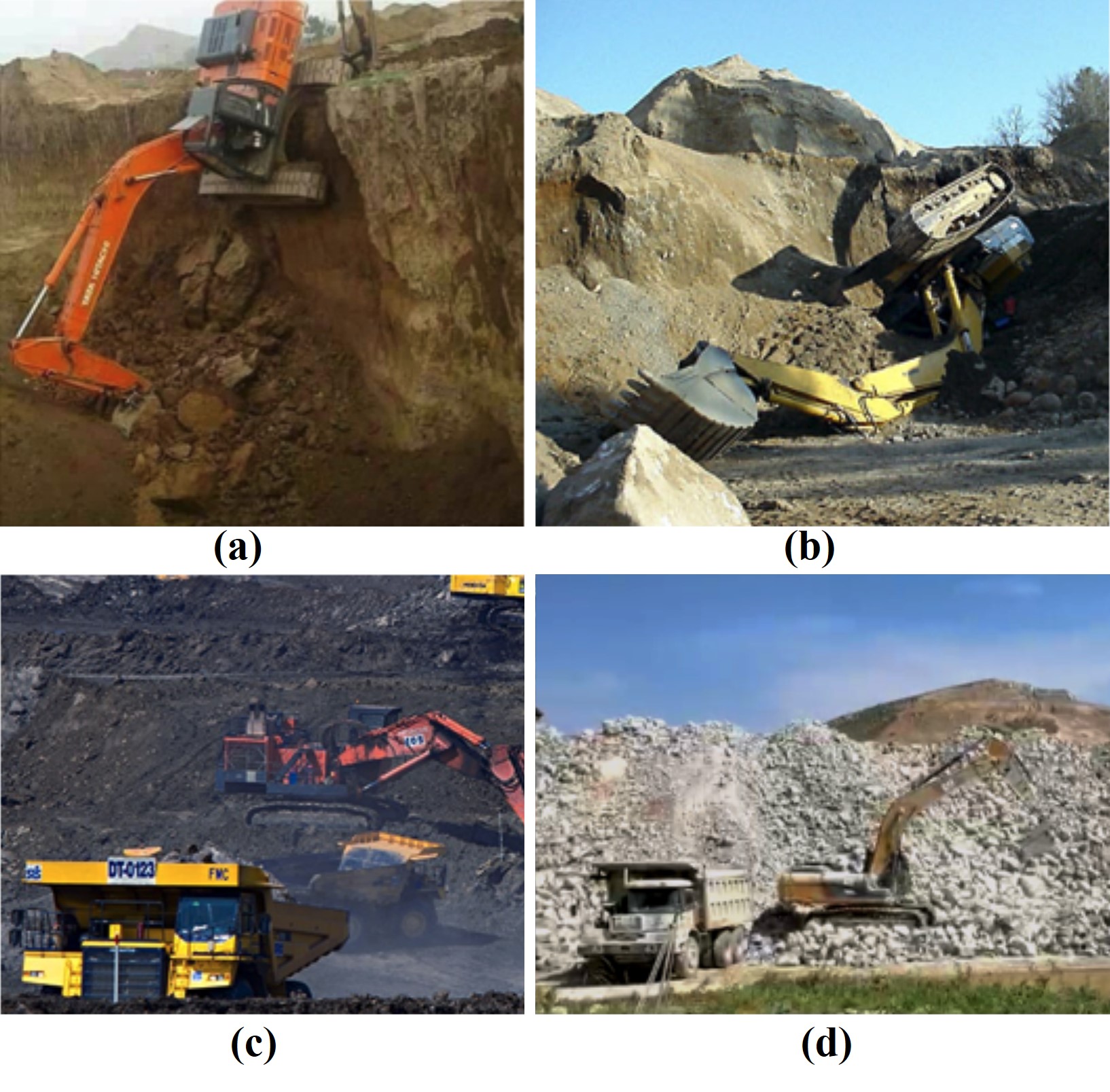}
    \caption{(a)(b) Land-falls in mining sites that cause excavator accidents, resulting in casualties or injuries of human operators. (c)(d) Excavators operate at extreme working conditions, such as remote area, high and low temperatures and exceeding dusty. Our AES system is designed to operate without human operators in such challenging scenarios.}
    \label{fig:excavator_accidents}
\end{figure}

 Excavation is the process of moving earth, rock, or other materials with tools, explosives or heavy  equipment. It is frequently used in different applications corresponding to mining, exploration, environmental restoration, archaeological sites, construction, etc. 
 In fact, excavators are considered as the most versatile heavy equipment with a huge market. In US, the excavator market size is about $5.25$ billion USD in 2008 \cite{exreport2018}. And currently China is the largest market for excavators.
In China, around 1.6 million  excavators were in operation during $2018$ and a total of $380$ thousand new excavators are projected to be sold in 2024~\cite{sohuindus}.
However, excavating is recognized as one of the most hazardous operations ~\cite{occupational1999excavation} and results in high number of injuries and deaths each year.
In USA, around $200$  casualties occur per year \cite{lew2002excavation}, caused by cave-in, ground collapse or other excavation incidents (Figure \ref{fig:excavator_accidents}).
The number of injuries and deaths would grow even
larger with more excavators being used. 




Currently, excavators are operated by human operators that have undergone specific training~\cite{afanuhpreventing}. This prolonged training includes not only learning excavating maneuvering techniques, but also safety regulations and standards.
In addition to the life-threatening incidents, human operators may also face extreme working conditions. 
For example, a mining site (Figure \ref{fig:excavator_accidents} (c) and (d)) is usually located at a remote area, or even inside a desert, where severe high or low temperatures can happen. Moreover, the remote location and long distances from cities results in limited availability of on-site excavator operators.  Hence, each worker suffers from prolonged working hours and loads, which can result in higher fatigue and more injuries~\cite{bauerle2018mineworker}. 
One solution is to develop autonomous excavators that can operate in such challenging and hazardous conditions without any human operators.  
An unmanned excavation system would largely reduce the number of casualties or injuries during excavation operation. Moreover, such an excavator could conduct tedious and repetitive tasks tirelessly for extended hours, thereby increasing the overall throughput.

Efficiency, robustness and generalizability are the three key requirements in terms of designing an autonomous excavator~\cite{dadhich2016key}. In other words, the autonomous excavator should be able to operate without human intervention, while finishing human-equivalent workload. Besides, the resulting excavator system should be capable of perceiving the surrounding environment, such as monitoring the target material status, detecting impurities and obstacles under extreme conditions, environment obstacles, and other uncertainties.

Current excavators can be classified based on their sizes.  A compact excavator normally weights less than $6$ tons, a standard excavator weighs between $7$ to $45$ tons, while a large excavator weights above $45$ tons and at most $900$ tons. Ideally, we want to develop autonomous systems for all these types of excavators.
The desired system should also be able to handle many types and sizes of materials, operate during raining or dust, and generate feasible motions while avoiding any collisions with obstacles. 

In terms of designing an autonomous excavator architecture that can operate robustly in real world scenarios, we address these challenges~\cite{hemami2009overview}: 
\begin{itemize}
    \item The system needs to operate under a large range of environment conditions that vary by the terrain types, weather, temperatures, lighting conditions etc. It is necessary that the perception module, which is responsible for understanding the surrounding environments, should be able to function in all such scenarios.
    \item Many times, an excavator is used to load a pile of material onto trucks. We refer to the material as the target material. Moreover, the shape of such target material pile changes after each loop of scooping  and dumping the material. As a result, we need real-time techniques for online modeling of the  shape of the pile;
    \item Scooping of material has to be successful regardless of the types or characteristics 
    of the materials, which vary based on the density, hardness or texture;
    \item After scooping the material, the excavator has to successfully dump the material into a truck, while avoiding any collisions with the truck, material pile or other obstacles in the environment; 
    \item The location of the dump truck body area
    could vary considerably during the excavation. We need to develop robust online detection methods to compute the truck pose.
\end{itemize}

\begin{figure}
    \centering
    \includegraphics[width=150mm]{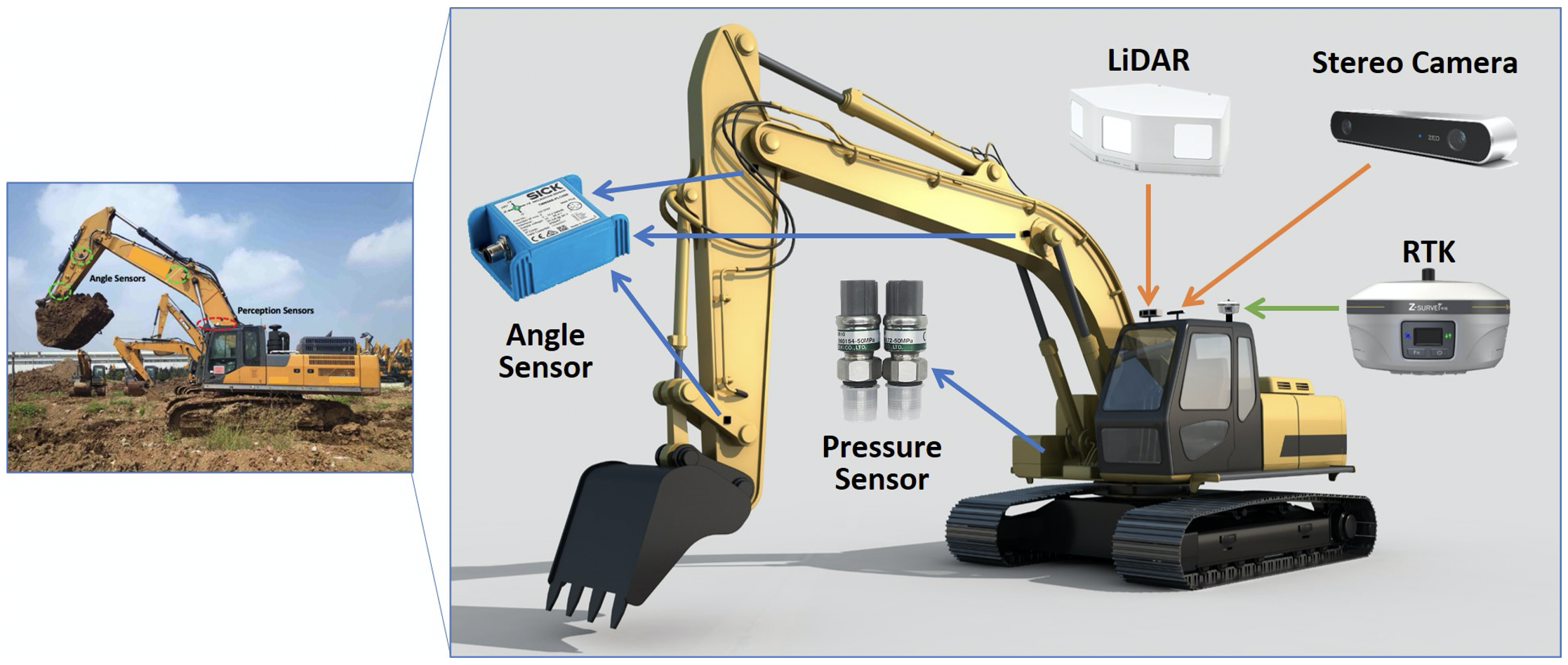}
    \caption{Our autonomous excavator system is equipped with state-of-the-art sensor hardware and drive-by-wire system. Our system is designed for material loading and dumping excavation task in various hazardous and difficult scenarios.  We highlight the various sensors used in our system (right) and being deployed on a real-world excavator (left). We evaluate the performance on compact and standard excavators.
    }
    \label{fig:layout_system}
\end{figure}

We present a novel system architecture and algorithms for autonomous excavator (AES) for material loading and soil excavation tasks. Our system integrates perception, planning and control  and is centered around the perception module. The planning and control modules are driven by  perception.  
We develop a perception system that fuses 3D-Lidar and camera outputs for perceiving 3D environments of source material piles and the target dumping area, detecting the material textures, impurities and blocking obstacles to improve the robustness of the whole system.
Moreover, the perception module detects unexpected objects, such as animals,
within the excavator operation zone and can trigger emergency stop to ensure the safety.
We also present a hierarchical planning module, which is composed of one target selection layer and one motion planning layer.
Our planning approach  combines inverse reinforcement learning (IRL) and imitation learning (IL), data-driven algorithms
with optimization-based methods.  
The target selection layer tries to learn the motion corresponding to excavation target selection, such as the next scooping position, 
from human operator demonstrations. The motion generation layer uncovers the motion pattern of human-operated excavation movements
offline and integrates this pattern with a stochastic optimization-based algorithm for trajectory generation of scooping and dumping motion.
Additional obstacle avoidance constraints, corresponding to trucks or buildings, are also added to our optimization formulation. 
 
\begin{figure}
    \centering
    \includegraphics[width=145mm]{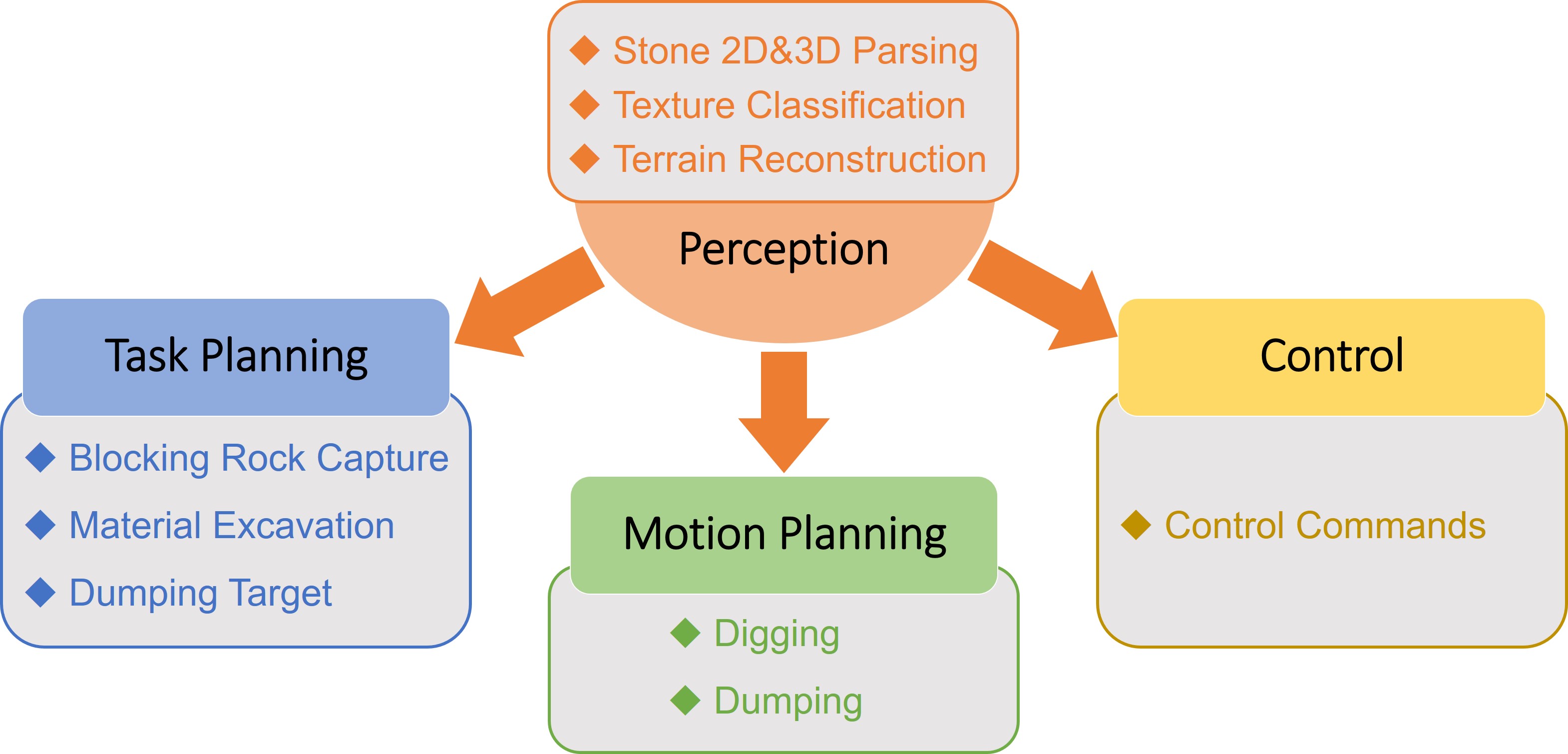}
    \caption{Overview of our Autonomous Excavator System (AES): the sensor data corresponding to Lidar and camera is captured by the perception module, 
    which parses the surrounding areas by checking for blocking obstacles, impurities of the material and material textures; This information is passed into the task planning module for determining the action corresponding to material or object capturing; The motion planning module generates a feasible trajectory based on these constraints; The controller follows the planned trajectory closely.  }
    \label{fig:system_overview}
\end{figure}

As shown in Figure \ref{fig:system_overview},
multiple perception algorithms taking inputs of camera images and Lidar point-clouds are integrated for different levels of perception. Such combination provides detailed scene understanding of the working zone, particularly the detailed condition of the target material. This is used to increase the robustness of the system and enables the system to operate without human operator assistance for prolonged hours.
Based on the rich information generated by perception module, the planning module takes advantage of data-driven and optimization-based methods to improve the efficiency.
Our system architecture with closely-combined perception and planning modules results in efficient excavation in different environments.
 
We use our modular architecture (Figure \ref{fig:system_overview}) to generalize the autonomous excavator system  for compact, standard and large excavators. 
This generalization is achieved by parameterizing the size of the excavators within the perception, planning and control modules of the system. 
Our autonomous excavator system is extensively tested and evaluated under controlled real-world scenarios, inside a designated test facility.
The details of various scenarios are highlighted in Table \ref{table:scenarios}.
During material loading excavation operation, the autonomous excavator performs the normal scooping motion and dumps the target material. At the same time, the autonomous system
handles terrain manipulation, obstacle avoidance and any water that may appear in the scene. In this context, terrain manipulation indicates that some objects like rocks or impurities could appear in the working area and block the normal excavation motion. In these scenarios, the excavator needs to lift and remove such kinds of objects, in order to successfully load the target material. Obstacle avoidance is performed during the loading operation and ensures  that the excavator should not collide with any loading trucks, the material pile after scooping as well as any non-targeted materials such as removed impurities. We highlight these tasks being performed in our test scenarios.

\begin{table}[h]
\caption{Scenarios Setup: We evaluated the performance of AES on  $8$ different scenarios. Our system operates robustly in these scenarios. The first column of the table lists the names of each scenario, while the other columns indicate whether the corresponding tasks need to be conducted in the related scenario. ``Y'' means the task is conducted in the scenario, while ``N'' means the otherwise. Terrain manipulation, obstacle avoidance and water recognition are three common tasks an excavator needs to perform for successful loading operation.
} 
\centering
\begin{tabular}{|c|c|c|c|} \hline
\diagbox{Scenario}{Elements}  & \begin{tabular}{c}Terrain \\ Manipulation \end{tabular} & \begin{tabular}{c}Obstacle \\ Avoidance \end{tabular}& \begin{tabular}{c}Handling \\ water \end{tabular}\\
\hline
\begin{tabular}{c}Scenario 1: \\ Material loading and dumping \end{tabular}
& N & N & N \\
\hline
\begin{tabular}{c}Scenario 2: \\ Terrain manipulation \end{tabular}
& Y & N & N \\
\hline
\begin{tabular}{c}Scenario 3: \\ Obstacle avoidance \end{tabular} & N & Y & N \\
\hline
\begin{tabular}{c}Scenario 4: \\ Loading with rain\end{tabular} & N & N & Y \\
\hline
\begin{tabular}{c}Scenario 5: \\ Terrain manipulation \\ with obstacle avoidance  \end{tabular} & Y & Y & N \\
\hline
\begin{tabular}{c}Scenario 6: \\ Terrain manipulation \\ with water handling  \end{tabular} & Y & N & Y \\
\hline
\begin{tabular}{c}Scenario 7: \\ Obstacle avoidance \\ with water handling
\end{tabular} & N & Y & Y \\
\hline
\begin{tabular}{c}Scenario 8: \\ Full stack scenario  \end{tabular} & Y & Y & Y \\
\hline
\end{tabular}
\label{table:scenarios}
\end{table}


We validate AES on two realistic applications. This includes recycling pipeline of hazardous industrial solid waste material, produced by various industrial activities.  We demonstrate that our approach can handle industrial waste, such as gypsum, dirt, gravel, or chemicals. 
In the second application,
we use AES to automate the material loading and dumping. 
We have extensively tested AES on 
the waste material recycling pipeline for 24 hours continuous operations in an indoor setting (see Fig.~\ref{fig:waste_disposal}).
We also compare the efficiency of our autonomous excavator system with skillful human operators for mass excavation and truck loading tasks. In terms of overall throughput, AES efficiency is closely equivalent to that of an expert  human operator.

\begin{figure}
    \centering
    \includegraphics[width=\linewidth]{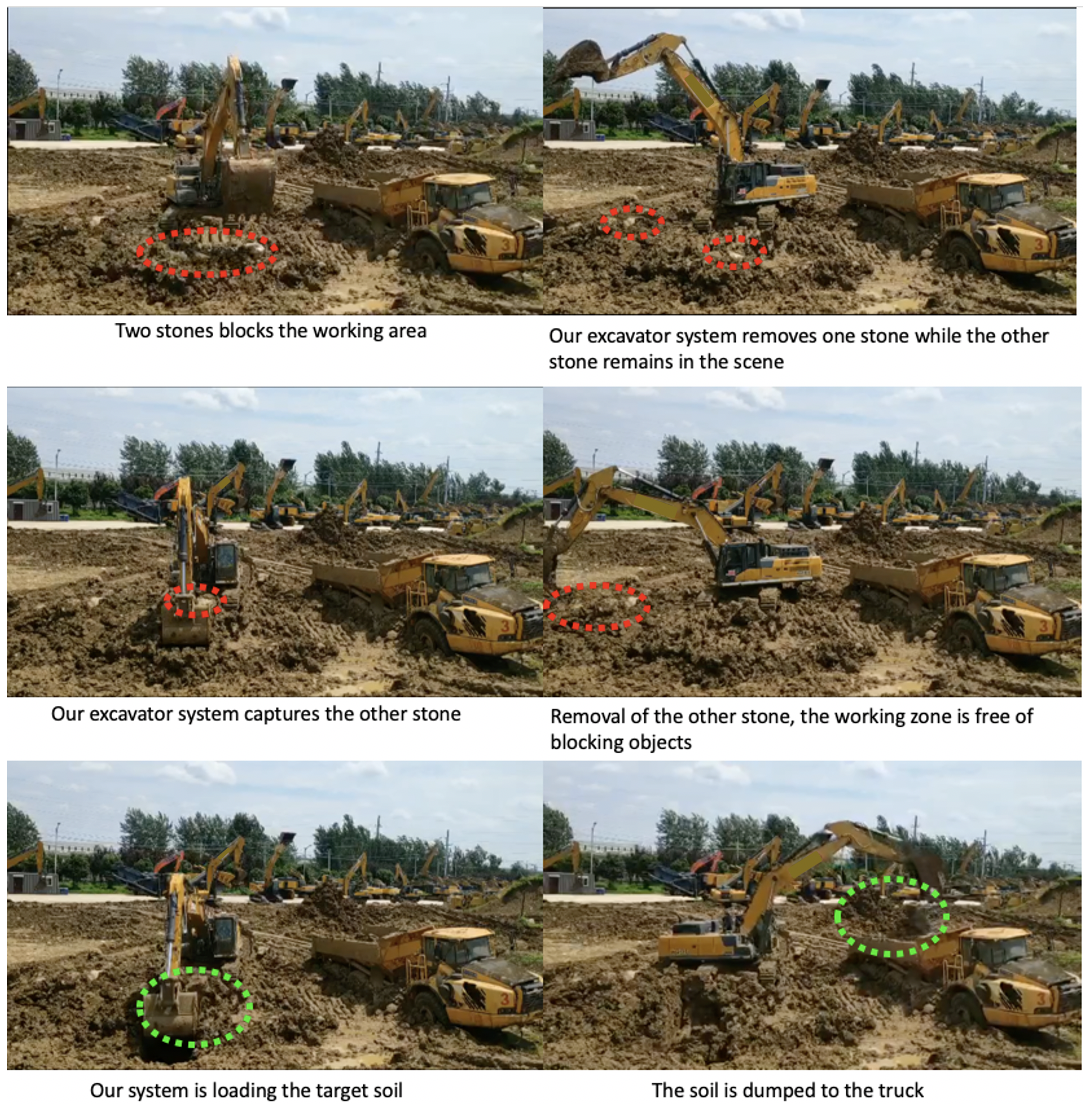}
    \caption{In this realistic mining site scenario, AES removes blocking obstacles if necessary and loads the material with success, even with water and impurities appearing in the scene. This highlights the robust capabilities in terms of handling such complex scenarios}
    \label{fig:combined_scene}
\end{figure}

We also evaluated AES for a mining application. Most mining sites are located at remote location and consist of  unconstrained environments, landfalls or cave-ins.
These hazardous scenarios require a robust perception module that can identify stones or rocks, along with appropriate motion planning algorithms that can be used to capture the rocks.  We highlight the performance on mining of handling blocking objects and obstacle avoidance in Scenario 2, Scenario 5, Scenario 6 and Scenario 8 (Figure \ref{fig:combined_scene}).

Overall, the novel components of our autonomous system (AES) include:

1) A perception-centered architecture that provides the excavator the capability of handling all the scenarios,
which results in high robustness of the excavator system in various dangerous and difficult indoor and outdoor environments.

2) A hierarchical planning module that combines data-driven method and optimization-based algorithms. 

3) Our autonomous excavator system is the first autonomous excavator that can continuously operate for $24$ hours without human assistance. We have
 thoroughly tested the performance of AES on multiple real-world scenarios. In addition, the system has been deployed on an application site and is being extensively tested by our partner. 

\subsection*{Previous Methods}

Several autonomous excavator related experimental systems and algorithms have been developed and tested.
Early development of autonomous excavators  dates back to $1990$s. In \cite{stentz98}, an autonomous loading system (ALS) is presented. 
ALS relies on a laser scan for perception to detect a single model of truck and the system handles single type of uniform soil without any impurity or change of texture.
Following this innovation, other system architectures for autonomous excavator have been proposed~\cite{kim2003framework,seo2011task,bradley1998development}.
For example, \cite{kim2003framework} introduces a conceptual framework of a semi-autonomous earthwork system, while human intervention is minimized.
Moreover, many heavy equipment manufacturers such as Caterpillar are also making effort to bring autonomy into construction industry~\cite{cateinnovation}. Their current focus is on remote control and semi-auto 

Along with the overall system integration, numerous methods and algorithms have been investigated for individual modules, especially planning and perception.
\cite{Singh1998MultiResolutionPF} introduces a classic planner architecture for earth-moving task. This architecture divides the planner into a coarse planner and a refined planner, which is responsible for planning the scooping region and excavation motion. Our autonomous excavator planning module uses a similar pipeline but improves each layer with improved planning algorithms for real-world deployment. 

As for coarse task planner, \cite{Seo2015TaskPD} focuses on the design of excavation motion sequencing problem to arrange an efficient schedule for the excavation task based on coverage planning rules.  
A control method is discussed in \cite{chang2002straight} to improve the motion trajectory tracking accuracy.
Recently \cite{Jud2017PlanningAC} propose an advanced planning and control algorithm to generate digging motions for excavation. This method is able to constrain the interaction force during the digging motion. 
These methods provides in-depth insights of autonomous excavator motion planning and control,
yet they focus on only limited scenarios and may not be able to handle complex real-world environments.

Recent advances in computer vision, deep learning and lidar sensors can be used to develop better perception capabilities for these excavators.  The robustness and accuracy of perception results is crucial for an autonomous excavator to understand the environment conditions and perform appropriate planning operations. 
A rock detection and segmentation method is discussed in \cite{dunlop2007multi}, based on the images of Mars terrain. Moreover, \cite{shariati2019towards} describes a safety check perception module to detect abnormally of an excavator bucket. 
These results focus on design and implementation of single perception functionality. We propose a perception module architecture that satisfies the overall fully automatic excavation perception requirements.

While some of these  prototype systems can perform certain tasks under restricted conditions,
rarely any autonomous excavators have been deployed in real-world scenarios. 
Our autonomous excavator (AES) is the first unmanned excavator system that has been deployed in real world scenarios and used by manufacturers of heavy machinery and the construction industry. We develop a novel perception system that fuses 3D-Lidar and camera outputs for perceiving the 3D environment. The combination of Lidar and camera helps detect both the type and location of the targets.
Our excavator planning module takes advantage of the advanced data-driven methods and stochastic optimization techniques to account for the uncertainties in sensed environment and the system itself. More importantly, our planning module is integrated and fully tested with our perception module. The combination results in a robust and efficient autonomous excavator that can operate for long hours without human intervention. 



\section*{Results}

\begin{figure}
    \centering
    \includegraphics[width=150mm]{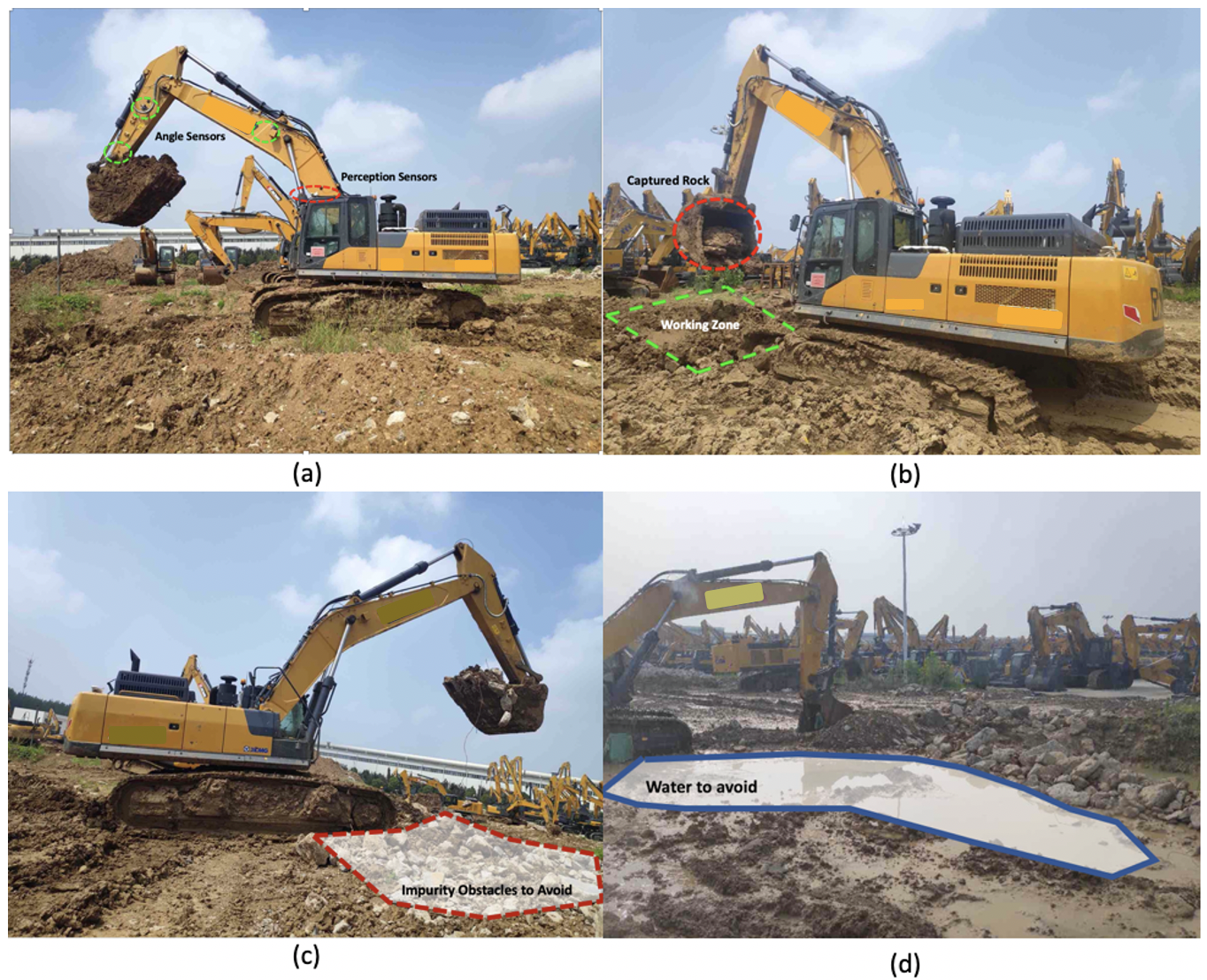}
    \caption{(a) Scenario 1: Our excavator system is operating autonomously in a basic scenario for excavator loading. In this scenario, the material soil is uniform. AES is able of handle the rough terrain and elevation. The perception sensors (red) and motion sensors (green)  are marked;
    (b) Scenario 2: Our autonomous excavator captures and removes the rock when it appears in the working zone and blocks the space for scooping the material.
    The working zone is the intersection of the location of the target material and the excavator range of motion;
    (c) Scenario 3: Our system successfully loads the material while avoiding the region with impurities. In case the bucket directly contacts the impenetrable impurity, the excavator arm will stop;
    (d) Scenario 4: In case a puddle or water appears in the scene, our perception module also segments these areas and feed their location to the planning module. In this case, our planner module generates the motion that avoids scooping into this area. 
    }
    \label{fig:4scenes}
\end{figure}

Our autonomous excavation system has been evaluated under multiple controlled real-world testing scenarios. 
Aiming to fully test the system capability, we set up $8$ scenarios in a closed test field, mimicking the common real-world use cases for an excavator. 

Based on the successful test results in these scenarios, we also evaluated the efficiency and robustness of the system in one of our deployment sites, a waste disposal factory. In this working scenario, we have shown that our AES system is able to continuously operate for $24$ hours without any human intervention. Meanwhile, our autonomous excavator system could achieve efficiency close to that of a human operator, measured by the amount of material handled during the same amount of time.
In addition, the excavator system is currently test and evaluated in a remote mining site, where the environment is substantially more hazardous and a human operator would confront many difficult conditions. 

\subsection*{System Verification Using Different Excavation Scenarios}

\subsubsection*{Scenario 1: Material loading and dumping}

This is a fundamental test scenario, where the material is free of impurities and there is no obstacle or water in the scene. 
This scenario helps us to evaluate terrain modeling perception and basic excavation motion generation for loading and dumping the target material into a dump truck (see  Figure \ref{fig:4scenes} (a)).

In this scenario, we further conduct comparison experiment to investigate the excavator efficiency between our autonomous system and human operator. 
In terms of average time used for
loading and dumping task, our autonomous system takes around $6 \%$ more than a human operator, while achieving larger bucket fill rate that corresponds to the percentage of the bucket that is loaded. 
Specifically, we divide each loading and dumping loop into four phases corresponding to ``scoop'', ``swing to truck'', ``dump'' and ``swing back''. 
``Scoop'' indicates the time taken to load the excavator bucket with target material; ``swing to truck'' indicates the time used for swinging the excavator bucket to align with the truck; ``dump'' means the time to unload the material from bucket to truck bed; ``swing back'' means the time used for swinging the bucket back to face the working area. The performance for each period is also provided in Table \ref{table:compare}.

\begin{table}
\centering
\caption{Efficiency comparison between our autonomous excavator and a human operator in terms of average time used for each loading loop and bucket fill rate. 
Our autonomous system takes $6\%$ more average time for a loading loop, while loading more amount of  material during each iteration measured by the bucket fill rate. Overall, the amount of material handled by AES per hour is closely equivalent to that of an experienced human operator.
}
\begin{tabular}{c||c|c|c|c|c|c}
\hline
 & \begin{tabular}[c]{@{}c@{}}Scoop\end{tabular} & \begin{tabular}[c]{@{}c@{}}Swing to truck\end{tabular} & Dump & \begin{tabular}[c]{@{}c@{}}Swing back\end{tabular} & Total & \begin{tabular}[c]{@{}c@{}}Bucket fill rate\end{tabular} \\ \hline
AES & 5.5 s & 8.5 s & 5.4 s & 5.7 s & 25.0 s & 105 \% \\
\hline
Human & 5.8 s & 6.2 s & 5.8 s & 5.7 s & 23.6 s & 86 \% \\
\hline
\end{tabular}
\label{table:compare}
\end{table}

\subsubsection*{Scenario 2: Terrain manipulation}

In this scenario (Figure \ref{fig:4scenes} (b)), the excavator has to lift stone or rock that blocks its working area for further excavation operation. The scenario is designed to extensively test the stone identification module and perform appropriate task/motion planning. The stone identification perception module outputs the 3D bounding boxes for the target stone or rocks, while the motion planning module generates a feasible trajectory for the excavator arm motion, in order to capture the object.


\subsubsection*{Scenario 3: Obstacle avoidance}
In this scenario, the detection of obstacles and collision-free trajectory generation are being evaluated,  as shown in Figure \ref{fig:4scenes} (c).
Moreover, the obstacles not only correspond to the loading trucks, surrounding buildings as well as material impurities and possibly the material that has been piled. The reason to avoid impurity is that these impenetrable impurities can jam or block the excavator arm motion. The reason to avoid the material pile is that the excavator arm should avoid any contact with the pile after filling up the bucket and before dumping the material into the truck. 


\subsubsection*{Scenario 4: Loading with rain}

In many circumstances, an excavator need to operate during rainy days. It is highly possible a puddle would appear around the working zone. In order to avoid scooping into the water, AES is capable of identifying the area filled with water. Based on these working conditions, we design a test scenario and verify the performance of our system, as shown in Figure \ref{fig:4scenes} (d).


\subsubsection*{Scenario 5: Combined scenario of terrain manipulation with obstacle avoidance}

This test scenario is closer to the realistic mining scenario, where Scenario 2 and Scenario 3 are combined. For robust mining operation, our perception module is designed to label the blocking objects and segment the obstacles. Our task planner first decides whether to remove the blocking objects and then the motion planner module generates the collision-free trajectory of the arm.


\subsubsection*{Scenario 6: Combined scenario of terrain manipulation with water handling}

When excavation happens during wet weather, the successful detection of both water and rocks is necessary. 
Hence we set up a new scenario that is a combination of Scenario 2 and Scenario 4.
In this case, the perception module reliably identifies the blocking objects as well as segments the water area.


\subsubsection*{Scenario 7: Combined scenario of obstacle avoidance with water handling}

In order to segment  the obstacles as well as the water area, our perception module takes advantage of the semantic segmentation method. During rainy days in scenarios with obstacles and impurities in the scene, we verify these segmentation capabilities of AES perception module.


\subsubsection*{Scenario 8: Full stack scenario}

This is the most difficult but realistic scenario, where all these challenges are combined. In this scenario, AES has to decide between manipulating the terrain and scooping the material, while avoiding the obstacles and any water area. We are able to evaluate the robust capabilities of our perception, planning, and control modules.

\subsection*{AES Deployment}

\subsubsection*{Waste Disposal Recycle}

\begin{figure}
    \centering
    \includegraphics[width=150mm]{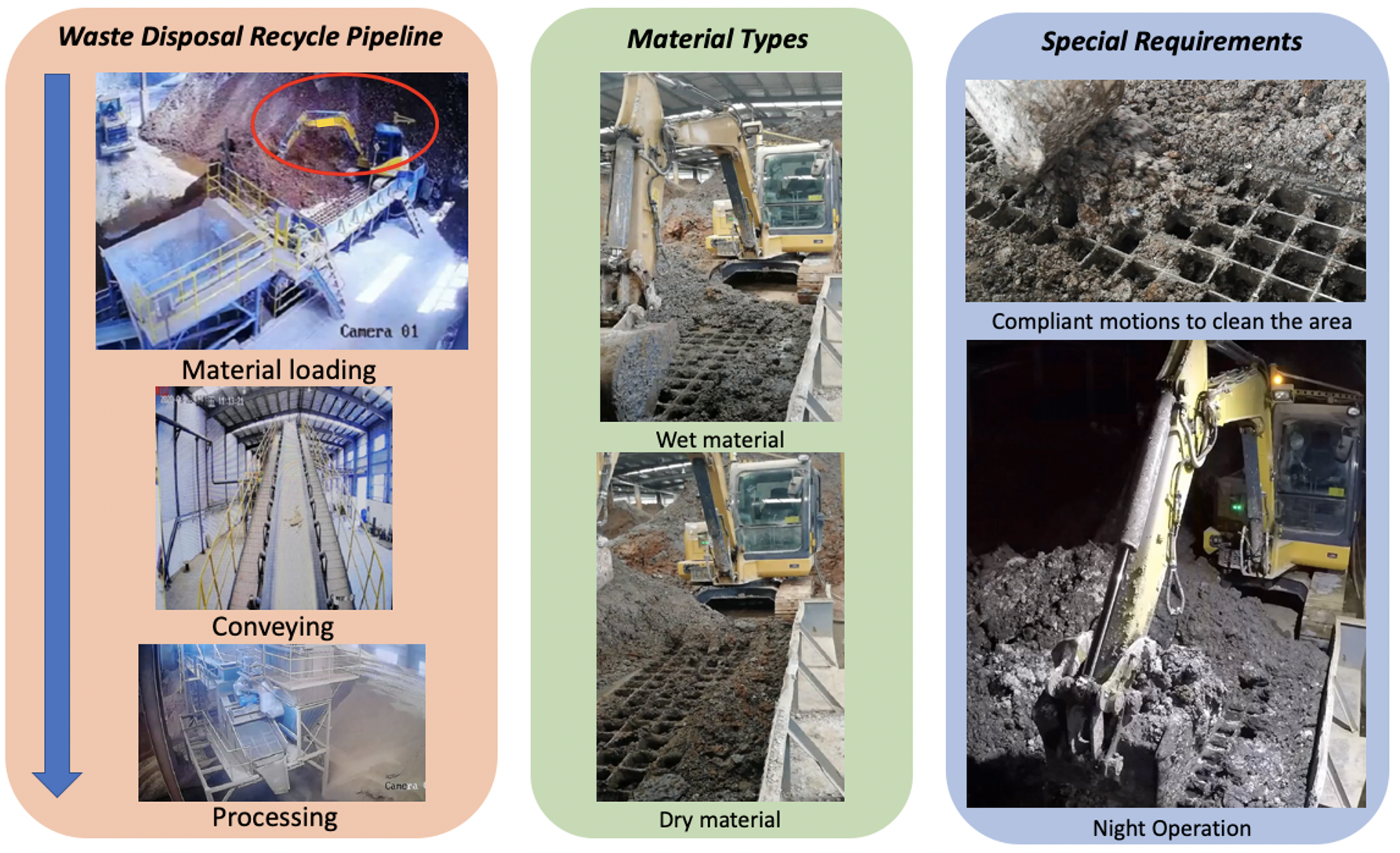}
    \caption{Robust and Prolonged Operation of AES:
    Excavator is used by the industry in waste recycle pipeline for loading various materials including toxic material for indoor dusty and smelly environments, which are harmful for human operators. In addition, the excavator sits next to a large pile of material, where landfall could happen. The pipeline needs to run 24 hours per day. Any malfunction of the excavator will stall the entire pipeline.
    During our stress test of the system, AES is able to operate a whole day of $24$ hours without any human intervention. Furthermore, the performance of AES is closely equivalent to human operators' performance.  We show the operation during daytime and nights, handling wet and dry materials.
    }
    \label{fig:waste_disposal}
\end{figure}

AES is the first autonomous excavator system used for such a complex, commercial application, and  can run continuously for $24$ hours without any human intervention.
In this application, the excavator is assigned to load the waste disposal material into a designated area. 
Afterwards, the material is transferred and recycled after loading. However, during the excavation process, the material could consist of excessive dust, which can be toxic to human beings. In addition, the material pile is not stable and could collapse, which is another threat to the human operators. The pipeline of the waste recycle is shown in Figure \ref{fig:waste_disposal}. The speed of material loading by the excavator has to coordinate with the speed of the belt conveyor and the material processing. Hence, there is a high efficiency requirement for our autonomous excavator. In addition to satisfying the efficiency requirement, our autonomous excavator system can handle both dry and wet material. Meanwhile, AES is able to function during night time. 

In this scenario, AES  is able to operate a whole day of $24$ hours without any human intervention. Meanwhile, AES can perform $145$ operations of excavation and dumping per hour; the amount of the handled material is as much as $36.25$ $m^3$ per hour for the $6.5$-ton excavator, which is closely equivalent to human operators' performance. Furthermore, AES performs consistently over the time, while the performance of human operators may have the variation.

\subsubsection*{Mining}

Another challenging scenario that is used to evaluate our system is mining site excavation (Figure \ref{fig:excavator_accidents} (d)). In this mining benchmark, the excavator loading process includes the following tasks: isolate the larger-sized rocks and remove them from the scene; load the smaller-sized rocks onto the loading truck. Our advanced perception module can reliably identify the pose and size of the rocks. Given these perception results, the motion planning module generates excavator arm trajectories corresponding to the rock captures. Our system is currently being used at the mining site.

\section*{Discussion and Conclusion}

Our newly designed autonomous excavator system is extensively tested for robustness and efficiency, and its effectiveness has been demonstrated above. 
Our system is the first ever autonomous excavator system put into deployment for prolonged operations. In the waste disposal application, our autonomous excavator system substitutes the human operator from the hazardous condition and achieves $24$ HPI (hours per intervention). Meanwhile, our system is currently used for a mining application, which is hazardous due to possible rock-falls. 



Compared with existing unmanned excavator system developed before, our overall architecture largely relies on the perception module for fine-grained understanding of the surrounding environment. Thus, our performance is mostly governed by the accuracy of perception algorithms and the sensor hardware. In heavy machinery applications, the reliability of the sensor hardware plays an important role. The recent developments in terms of better Lidar and camera sensors, along with improved computer vision algorithms, can further improve the performance of our perception module.
Moreover, we currently have not considered the scenarios where snow or ice appears. Such scenarios  introduce many challenges for not only the  perception module, but also  for the  planning and control modules. Our goal is to further evaluate the performance in more challenging weather and illumination conditions.


\section*{Methods and Materials}


\subsection*{Software Architecture}

There are four software modules in our system. The perception module takes care of sensing various obstacles, modeling the terrain and classifying the material, as well as locating the loading truck. The planning module generates the excavator arm motion according to the perception output. Meanwhile, the control module calculates the commands that sent to the excavator in order to track the desired motion. The application layer of the software adjusts the other modules based on the application. 

All modules run simultaneously as nodes under the ROS~\cite{ros} framework.
In this section, we provide more details on perception and planning modules, which are the key components that enable our system to be deployed in real-world scenarios. The overall architecture of AES  is shown in Figure \ref{fig:system_overview}.


\subsection*{Perception}



The perception module for an autonomous excavator focuses on understanding the condition of the unstructured working zones. More specifically, the perception module mainly needs to determine the status of the material, i.e.
\begin{itemize}
\item Identify the blocking obstacles that need to be captured and removed;
\item Detect the impenetrable portion of the material to avoid direct contact with the excavator arm;
\item Recognize the texture of the material and model the shape of the material pile to perform the loading operation.
\end{itemize}
We need to satisfy all these goals for different materials, under difference weather and lighting conditions.

Towards this end, our perception module uses an hierarchical approach for detection and identification of larger-sized objects, small-sized targets and material textures.
Our perception module exploits extensively on state-of-the art algorithms like semantic segmentation~\cite{chen2017deeplab}\cite{liu2019auto} and super pixel \cite{he2017mask}.

\subsubsection*{Blocking Objects Detection}

In many excavation scenarios, blocking objects indicate the large-sized rocks or accumulated undesired material impurities that can usually cause the failure of the excavation motion. Human operators are able to remove these kind of objects after certain amount training. In order to robustly remain in autonomous operation without human assistance, AES needs to have equivalent ability of detecting these objects. This part shows the results corresponding to blocking obstacle detection, where the pose and size of an obstacle are obtained through RGBD camera and Lidar. The information is fed to the planning module to compute the trajectory to perform the task of rock-removing movement.

We employ the Mask R-CNN~\cite{he2017mask} algorithm to identify the blocking obstacles in the scene with RGB input. Based on the segmentation results, the depth information from camera and Lidar are combined and used for fitting the 3D bounding boxed to each  detected obstacles. An example of stone pose estimation
is shown in Figure \ref{fig:impurity_seg}. For training the Mask r-cnn model, the training data-set contains $4000$ rock or stone instances. The training period took $2$ hours using $4$ Nvidia 1080 GPU.

\begin{figure}
    \centering
    \includegraphics[width=150mm]{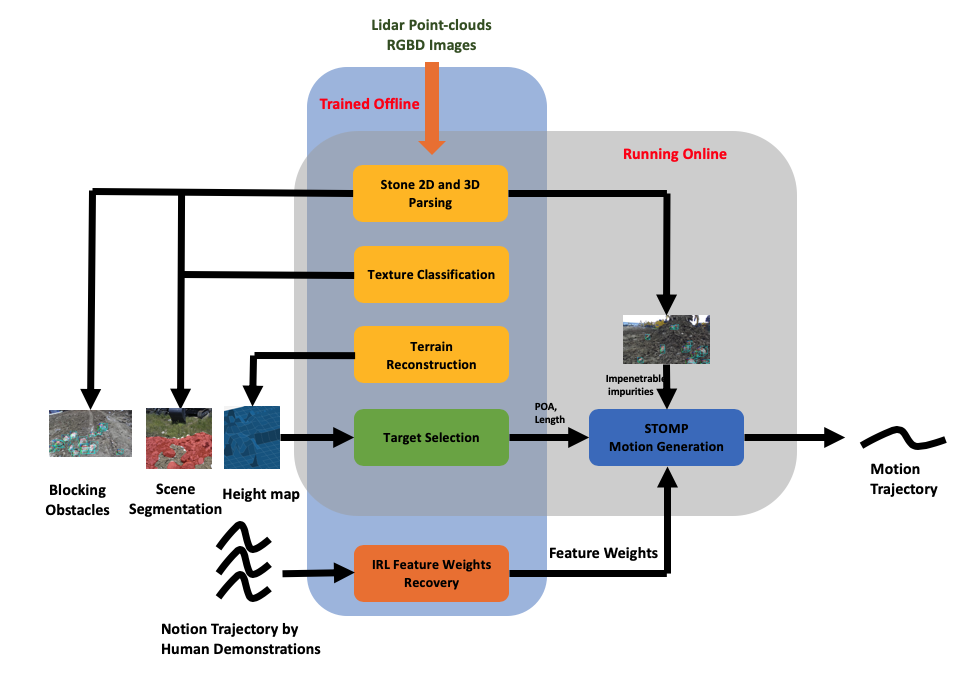}
    \caption{AES architecture: The perception module provides different layers of results to task planning and motion planning module. This perception-in-center layout helps provide rich information of the surrounding environment. 
    }
    \label{fig:block_diagram}
\end{figure}

\subsubsection*{Obstacle Identification}

\begin{figure}
    \centering
     \includegraphics[width=0.95\linewidth]{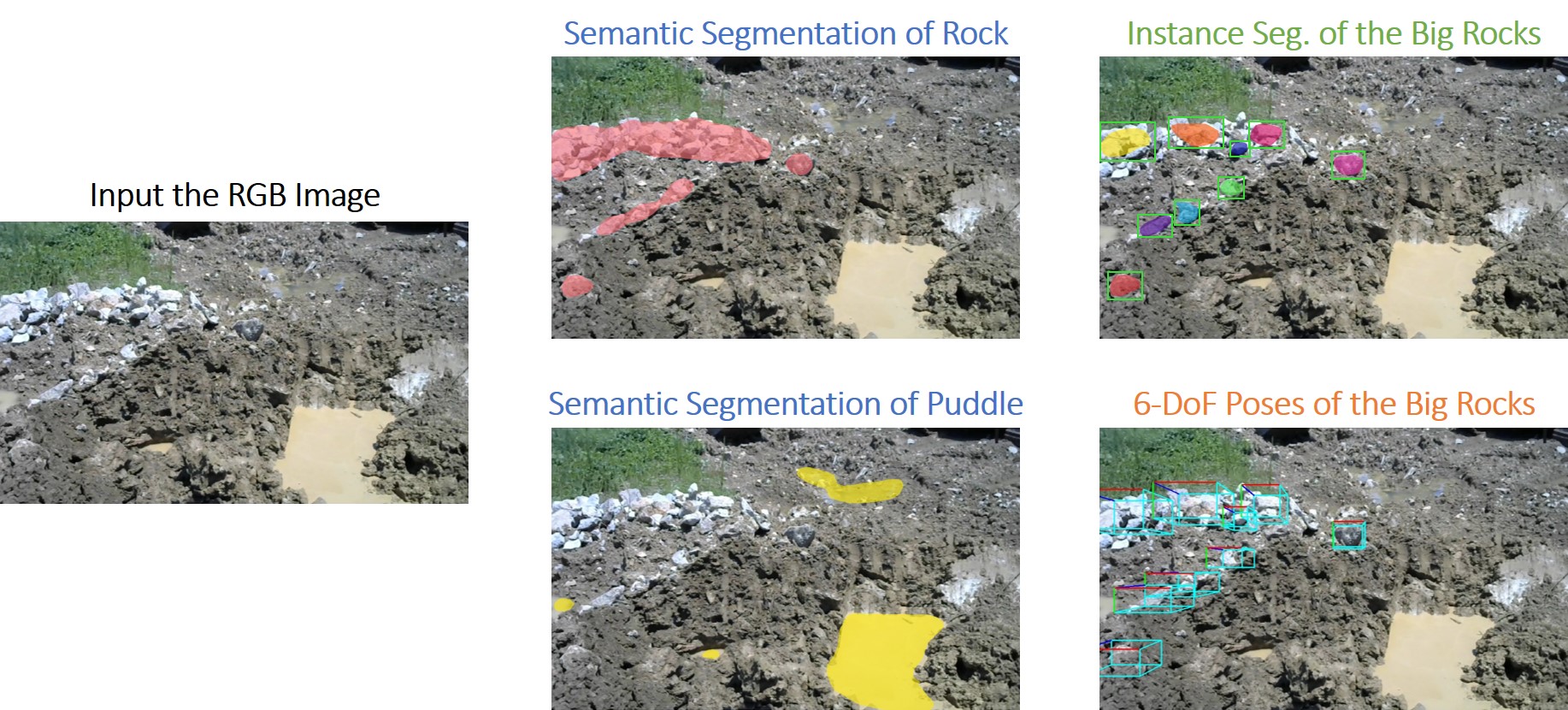}
    \caption{Obstacles Segmentation: Obstacles such as impurities, stone and water 
    need to be avoided during the excavation operation. 
    With obstacles being labeled through semantic segmentation, our autonomous excavator system is able to avoid any direct contacts with these obstacles. 
    Our perception exploits Mask R-CNN model for semantic segmentation from RGB images, the segmentation results are combined with depth information to generate the bounding boxes.
    }
    \label{fig:impurity_seg}
\end{figure}

Another difficulty that commonly happens during the excavation operation 
is to avoid collisions with obstacles. Such obstacles include the material impurities, the loading truck, the material pile after the scooping operation, etc.
When the material contains hard impurities, they will prevent the further movement of the excavator arm after direct contact. This is more likely to happen when the end-tip or the end-effector of the excavator bucket contacts such impurities. In this case, human operator normally performs reasonable choice in terms of excavation motion that can avoid direct contact between the impurities and the end-tip of the bucket. 
TTo identify such scenarios, we perform semantic segmentation as shown in Figure \ref{fig:impurity_seg}.

\subsubsection*{Texture Classification}

The properties of the loading material such as the density and hardness could largely affect the excavation motion. We employ a visual texture classifier based on the RGB image inputs to predict the type of the material. 
In specific, to well train the texture classifier, we collect a texture dataset which are captured from two environments: wild environment and chemical wastes disposal. The dataset contains 8 classes, named, wild environment: dry soil, wet soil, stone and mud stone and chemical wastes: phosphorus powder, phosphorus lump, titanium and mixture of titanium and phosphorus. The dataset contains $7563$ images, and $6720$ images for training and $843$ images for testing. We employ Deep Encoding Pooling Network~\cite{Xue_2018_CVPR} with attention strategy for texture classification, and accuracy of $100\%$ can be achieved. The offline training period takes 1 hour using an  Nividia $V100$ GPU.

\subsubsection*{Loading Truck Pose Estimation}

As for truck pose estimation, a template truck model with 3D point cloud representation is computed offline. This is performed by scanning the truck from multiple perspectives and recording the point clouds. Next, we segment each point cloud that corresponds to the truck and compose the resulting point clouds together to obtain a truck template. During online truck pose estimation, the truck template model is matched with the point cloud from Lidar scan using iterative closest point (ICP) based algorithm. Finally, the estimated truck pose is used to determine the bucket location for material dumping.

\subsection*{Planning}

Our planning module is closely related to perception module, as shown in Figure \ref{fig:block_diagram}.
To handle the real-world material loading task, the planning module needs to explicitly consider the terrain shape as well as the positions of the obstacles. 
Our hierarchical planning module automatically selects the target of excavation based on the perception results. The detailed arm motion is  generated based on this target. Thus, the planning module is composed of a task planning algorithm and a motion generation algorithm. This de-coupling structure allows us to combine data-driven method with an optimization-based algorithm. 

\subsubsection*{Autonomous Excavation Target Selection}
We use a data-driven method for the target selection module. 
This module aims to learn the routine of a human operator according to the terrain observation of the excavation working zone. 
In this application, the terrain observation is selected as the 2.5D grid-based height map \cite{gu2008rapid} of the working area. The predicted output are the location of point of attack (POA) and bucket travel length, where POA represents the point that excavator bucket first contacts the material, 
the travel length means the distance the bucket travels before lifting the material. A neural network model is designed for the learning task and is described as
\begin{align*}
    y = f_{core}(x), ~~~ \theta = y^T M, ~~~ z_x = f_{lon}(\theta M), ~~~ z_y = f_{lat}(\theta M), ~~~ z_l = f_{l}(\theta M),
\end{align*}
where $x$ is the observation, $f_{core}$, $f_{lon}$, $f_{lat}$ and $f_l$ are multilayer perceptron (MLP) with trainable parameters, $M$ is a matrix with trainable elements, $z_x$ and $z_y$ are the longitudinal, lateral coordinates of the POA, $z_l$ is the length that the excavator bucket travels before lifting.
This model follows the idea of neural programming interpreter \cite{reed2015neural}.

\begin{figure*}[h]
    \centering
    \includegraphics[width=0.98\linewidth]{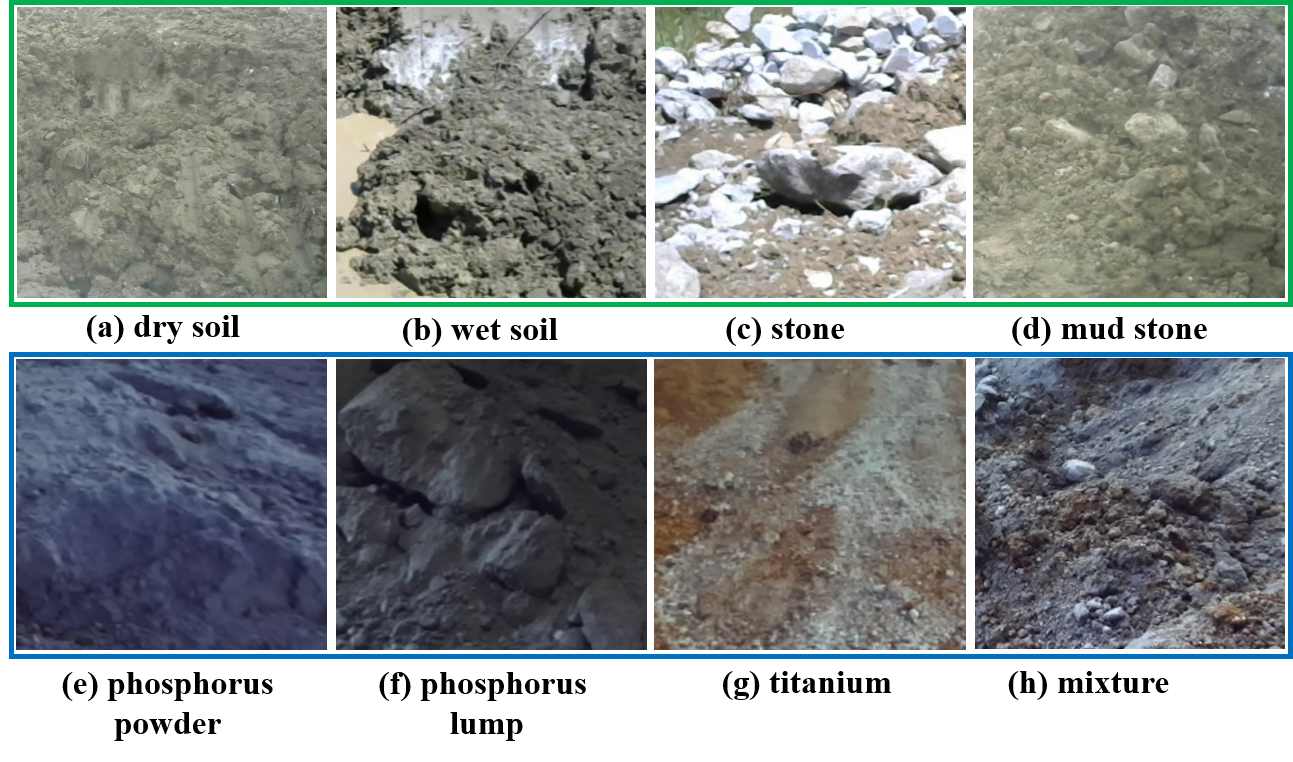}
    \caption{The figure shows the examples of the data-set used for  texture classification. Images in the top row correspond to  textures in the wild environment, while images in the bottom row represent mean different type of chemical wastes. 
    We trained a classification model for identify the class labels for each material type. 
    }
    \label{fig:texture}
\end{figure*}

\subsubsection*{Excavation Motion Generation}

This motion generation module outputs the trajectory in the excavator joint space based on the target selection module. For this part, we leverage a data-driven method to uncover the human operated excavation movement pattern. The pattern is later implemented into a optimization-based method for generating the motion trajectory. 

We use inverse reinforcement learning (IRL) \cite{kalakrishnan2013learning} algorithm to uncover the motion pattern. 
Given a trajectory $\tau$, the cumulative cost $C(\tau)$ is defined as
\[
C(\tau) = w^T \Psi(\tau),
\]
where $\Psi$ is a user defined feature function and $w$ is the weight vector associated with the feature. 
Given multiple collected human operated excavation motion trajectories, the target is to learn the feature weight vector $w$, which can be obtained by solving
a convex optimization problem
\begin{align*}
    minimize_{w} \sum_{i=1}^D \log \sum_{k=1}^K e^{-w^T (\Psi_{i, k} - \Psi_i^*)},
\end{align*}
where $K$ is the number of sampling points of one trajectory and $D$ is the number of demonstrated trajectories.
Based on the determined POA and bucket travel length, the start and end excavator arm configuration can be calculated through inverse kinematics. The features we define are the squared error between certain configuration and the end configuration w.r.t. each joint, as well as the changing rates of the errors. This learning process is done offline.

During the online motion generation, the concluded feature weights and cost features are used in a stochastic trajectory optimization for motion planning (STOMP) \cite{kalakrishnan2011stomp}. In addition to the human pattern, STOMP allows us to consider the obstacles in the scene as well. Recall that there always exist impenetrable impurities in the material and our perception module is able to identify the locations of such impurities. Therefore, additional cost features related to these obstacles are added to the overall cost function is the STOMP algorithm. 


\subsection*{System Hardware}

In this application, the robot platform (Figure \ref{fig:layout_system}) are hydraulic excavators equipped with drive-by-wire system. 
Currently we have developed and tested on two different sizes of excavators, one is a $6.5$-ton compact excavator and one is a $49$-ton standard excavator.
These excavation platforms offer enormous output power to successfully conduct various excavation tasks. The manufacturer provides a control interface through CAN bus, so the entire unit can be controlled by software. To ensure safety, a fall-back human control mechanism is implemented in case of an emergency. 

To sense the excavator locations and motions, multiple sensors are installed. We use a Huace real-time kinematic (RTK) positioning device for providing the location of the excavator.
Honeywell inclinometers are used for measuring the angles of different joints of the excavator. During the hardware tests, such sensors have shown relatively accurate readings amid the excessive hardware vibrations. 
A combination of Livox mid-100 light detection and ranging (LIDAR) sensors and ZED RGBD cameras collect the environmental information for the perception module to process and analyze for understanding the surroundings. 

One Intel i7 computer with $8$ CPUs and $2$ Nvidia GPUs, running Linux system hosts the perception, planning and control sofrware modules. 
An industrial control board based on STM32F407 arm micro-controller is used for low-level communication between the computing unit and the excavator.


\clearpage


\bibliographystyle{unsrt}
\bibliography{scibib}

\end{document}